# Contractive De-noising Auto-encoder

Fu-qiang Chen, Yan Wu[*], Guo-dong Zhao, Jun-ming Zhang, Ming Zhu, Jing Bai

College of Electronics and Information Engineering, Tongji University, Shanghai, China

**Abstract.** Auto-encoder is a special kind of neural network based on reconstruction. De-noising auto-encoder (DAE) is an improved auto-encoder which is robust to the input by corrupting the original data first and then reconstructing the original input by minimizing the reconstruction error function. And contractive auto-encoder (CAE) is another kind of improved auto-encoder to learn robust feature by introducing the Frobenius norm of the Jacobean matrix of the learned feature with respect to the original input. In this paper, we combine de-noising auto-encoder and contractive auto-encoder, and propose another improved auto-encoder, contractive de-noising auto-encoder (CDAE), which is robust to both the original input and the learned feature. We stack CDAE to extract more abstract features and apply SVM for classification. The experiment result on benchmark dataset MNIST shows that our proposed CDAE performed better than both DAE and CAE, proving the effective of our method.

## 1 Introduction

Generic neural network is composed of three layers, one input layer, one hidden layer and one output layer. Generally, for the input layer, we need only input the original data to it. While for the hidden layer, we need to compute a kind of transformation of the input to get a kind of feature from the input by a special kind of feature mapping function. And for the output layer, we can get the values of the neurons in the layer as the label of the corresponding input also by a special kind of feature mapping function. In 1988, Bourlard and Kamp [1] first proposed auto-association, a kind of neural network which replaces the output layer with the data same as the input layer, while in the generic neural network the output layer is set with the label of the corresponding input sample. And auto-association is the original auto-encoder. It can be learned by error back propagation (BP) [2]. In 2006, Hinton and Salakhutdinov [3] stacked several layers of auto-encoders to get deep auto-encoders for dimension reduction, and they pre-trained the deep auto-encoder with restricted Boltzmann machine (RBM) [4], which is a special kind of neural network based on Boltzmann machine. Here the restricted is referred to that there are no intra-layer connections both in the visible layer and the hidden layer, and there are only inter-layer connections between the neurons in the visible layer and the hidden layer.

---

[*] Corresponding author. Email: yanwu@tongji.edu.cn

In 2008, Vicent P. et al. [5] proposed de-noising auto-encoder (DAE), a kind of improved auto-encoder which first corrupts the original data and then transforms or encodes the corrupted input to the hidden layer and then decodes to the 'corrupted'[1] output layer, where the 'real' output layer is identical to the original input layer. Then by error back propagation, de-noising auto-encoder is robust to the original input.

In 2010, Vicent P. et al. stacked de-noising auto-encoder to form deep neural network to learn useful representations [6], and they showed that SDAE performed better than deep belief nets (DBN) in several cases. Besides, they found DAE could learn Gabor-like edge detectors from natural image patches and larger stroke detectors from digit images [6].

Then in 2011, Rifai et al. [7] proposed another kind of improved auto-encoder, contractive auto-encoder (CAE). They improved auto-encoder by introducing the Frobenius norm of the Jacobean matrix of all the learned features in the hidden layer with respect to the original input for the reconstruction loss function in AE. CAE can be regarded as an approach for nonlinear dimension reduction related to manifold learning. And they found that CAE outperformed DAE on some datasets.

For DAE, it only takes the corruption of the original data into consideration, which makes the auto-encoder robust to the input. While for CAE, it adds a penalty term to the reconstruction function on the basis of the original auto-encoder. And it makes the learned feature robust.

In this paper, we combined DAE and CAE, and we proposed contractive de-noising auto-encoder (CDAE).

---

[1] Here, we call the output layer 'corrupted' because the output layer is got based on the 'corrupted' input.

CDAE is both robust to the input and also makes the learned feature robust by adding a penalty term to the reconstruction loss function. After extracting feature with CDAE, we apply SVM for classification. And our experiment on benchmark dataset MNIST shows that our method surpasses both CAE and DAE, which proves the effective of our method.

## 2   CDAE

In this section, we will present our proposed algorithm, CDAE. Firstly, we will introduce the formulation of generic auto-encoder and its two variants, i.e., DAE and CAE.

### 2.1   AE

Let **x** ( in this paper, **x** is a column vector unless otherwise stated ) represents a specific input or sample for AE, $d_v$ the input layer size or the dimension for the input, $d_h$ the size of the hidden layer or the number of neurons in the hidden layer.
To encode **x** into the hidden layer, we can use the following activation function:

$$\mathbf{h} = sigmoid(\boldsymbol{W}\mathbf{x} + \mathbf{b}), \qquad (1)$$

where $sigmoid(x) = \frac{1}{1+e^{-x}}$.

Another commonly used activation function is

$$tanh(x) = \frac{e^x - e^{-x}}{e^x + e^{-x}}.$$

In formula (1), $\boldsymbol{W}$ is a matrix of $d_\mathbf{h} \times d_\mathbf{x}$ (usually $d_\mathbf{h} < d_\mathbf{x}$), while **b** is a $d_\mathbf{h} \times 1$ bias vector for the hidden layer. Thence, $\boldsymbol{W}\mathbf{x}+\mathbf{b}$ is a column vector, and the sigmoid function maps $\boldsymbol{W}\mathbf{x}+\mathbf{b}$ element-wisely, then we can get a column vector of size $d_\mathbf{x} \times 1$. Then we need to decode the hidden layer to the output layer according to the following formula:

$$\mathbf{x}_{rec} = sigmoid(\boldsymbol{W}^T\mathbf{h} + \mathbf{c}),$$

where the subscript 'rec' means reconstruction considering that the output of an auto-encoder is the input itself. And **c** is a $d_\mathbf{x} \times 1$ bias vector for the output layer.

Last, we need to compute and minimize the loss function or the reconstruction cost function with respect to $\boldsymbol{W}$, **b** and **c** by error back propagation as following:

$$\text{Minimize } L_{AE} = \frac{1}{N}\sum_{\mathbf{x}\in D}(\mathbf{x}-\mathbf{x}_{rec})^2 \quad (2)$$

where ***D*** is the set composed of all the input samples, while ***N*** is the cardinality[2] of set ***D***. And here the loss function is the squared error function, which is a widely used cost function in neural network. Another form of cost function (cross-entropy loss function) is as follows:

$$L(\mathbf{x},\mathbf{x}_{rec}) = L_{cor-en} = -\sum\left(\mathbf{x}_i log\mathbf{x}_{rec_i} + (1-\mathbf{x}_i)log(1-\mathbf{x}_i)\right).$$

For clarity, we show the architecture of the auto-encoder in Fig. 1.

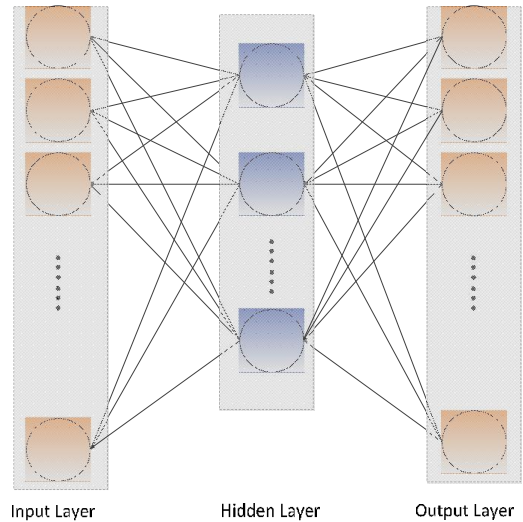

**Fig. 1.** The architecture of Auto-encoder, in which the output is identical with the input. And this is the main difference of Auto-encoder from the generic neural network, whose output layer is the label of the corresponding input.

### 2.2  DAE

In DAE, we need to corrupt the original data firstly. There are two common ways to corrupt the original data, additive isotropic Gaussian noise and binary masking noise.

---

[2] For finite set, the cardinality of a set is the number of the elements in the set. For detail, the readers are referred to http://en.wikipedia.org/wiki/Cardinality

Here, we denote the corrupted version of the original input sample **x** as $\tilde{\mathbf{x}}$. Then, for additive isotropic[3] Gaussian noise,

$$\tilde{\mathbf{x}} = \mathbf{x} + \mathbf{N}(\mathbf{0}, \sigma^2 \mathbf{I}),$$

where $\sigma$ is a small constant maybe usually smaller than 0.5 [6]. And **I** is the identity matrix, i.e., with all the elements in the diagonal equal **1** while other elements all equal **0**. And for binary masking noise, a small fraction of the input sample **x** is set to be zero, and the fraction can take values on 10%, 25% etc.

For a specific input **x**, first we need to get the corrupted version of **x**, $\tilde{\mathbf{x}}$. Then we can get

$$\tilde{\mathbf{h}} = sigmoid(\mathbf{W}\tilde{\mathbf{x}} + \mathbf{b}),$$

**W** and **b** satisfies the same condition as in AE.

Subsequently, we can get the reconstructed version of input **x** by

$$\tilde{\mathbf{x}}_{rec} = sigmoid(\mathbf{W}^T\tilde{\mathbf{h}} + \mathbf{c}).$$

Finally, we can get the object function of DAE as following:

Minimize $L_{DAE} = \frac{1}{N}\sum_{\mathbf{x} \in D}(\mathbf{x} - \tilde{\mathbf{x}}_{rec})^2$ (3)

For clarity, we show the architecture of de-noising auto-encoder in Fig. 2.

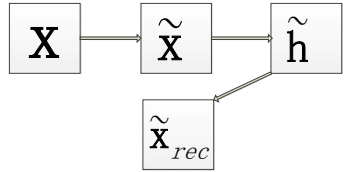

**Fig. 2.** The architecture of de-noising auto-encoder.

### 2.3 CAE

We can get CAE by adding one kind of penalty term to the original cost function (i.e. (2)) of AE. The added term is the Frobenius norm[4] of the Jacobean matrix of the

---

[3] Here, the 'isotropic' means that the variances of all the input dimensions are equal.

[4] For detail of Frobenius norm, the readers are referred to

learned feature in the hidden layer with respect to the original input. The corresponding formula can be shown as following:

Minimize $L_{CAE} = \frac{1}{N}\sum_{\mathbf{x}\in D}((\mathbf{x} - \mathbf{x}_{rec})^2 + \lambda \|J_h(x)\|_F^2)$. (4)

In formula (4), the $\lambda$ is a parameter and it takes values of 0.1 in this paper [7]. For activation function $sigmoid$, the second term in formula (4) can be calculated as following:

$$\|J_h(x)\|_F^2 = \sum_{i=1}^{d_\mathbf{h}}\left(\mathbf{h}_i(1-\mathbf{h}_i)\right)^2 \sum_{j=1}^{d_\mathbf{v}} W_{ij}^2,$$

for activation function $tanh$, the second term in formula (4) can be calculated as following:

$$\|J_h(x)\|_F^2 = \sum_{i=1}^{d_\mathbf{h}}\left((1+\mathbf{h}_i)(1-\mathbf{h}_i)\right)^2 \sum_{j=1}^{d_\mathbf{v}} W_{ij}^2.$$

## 2.4 CDAE

From the above formula (2-4), we can see that DAE improves the traditional AE by corruption-reconstruction (first corrupt the original data and then reconstruct the original data by minimize the reconstruction loss function), which makes auto- encoder robust to the input.

While CAE improves the traditional AE by introducing a penalty term in the object function for every sample, the Frobenius norm of the Jacobin matrix of the learned hidden feature with respect to the original input. CAE makes AE robust by changing the object function, which indeed makes some neurons in the hidden layer less active to learn more helpful and robust feature.

To combine the advantages of both CAE and DAE, we propose another improved AE. Here we call it contractive de-noising auto-encoder (CDAE). The object function of CDAE can be written as following:

$L_{DAE} = \frac{1}{N}\sum_{\mathbf{x}\in D}((\mathbf{x} - \mathbf{x}_{rec})^2 + \lambda \|J_{\tilde{\mathbf{h}}}(\tilde{\mathbf{x}})\|_F^2)$. (5)

---

http://mathworld.wolfram.com/FrobeniusNorm.html

We first corrupt the original data **x** to get $\tilde{\mathbf{x}}$. Then we can map $\tilde{\mathbf{x}}$ to the hidden layer and we can get $\tilde{\mathbf{h}}$.

Subsequently, we can get the reconstruction of the original sample $\mathbf{x}_{rec}$ by decoding $\tilde{\mathbf{h}}$. From the object function, we can see that CDAE is a directly improved version of CAE, by adding the corruption and de-noising. And similarly, CDAE is also a directly improved version of DAE, by adding a penalty term, the Forbenius norm of the learned feature with respect to the corrupted version of the original input, to the object function.

## 3      Experiment

Considering that we'll apply support vector machine (SVM) [8] in our experiment, we'll introduce SVM briefly in this section. SVM is a commonly adopted classifier in various applications. It is based on kernel function, dual problem and convex quadratic programming. For any two specific inputs, **x** and **y,** there are three common kernel functions in support vector machine as following:

$$K(\mathbf{x},\mathbf{y}) = (\mathbf{x} \cdot \mathbf{y} + 1)^p, \qquad (6)$$

$$K(\mathbf{x},\mathbf{y}) = e^{-\frac{\|\mathbf{x}-\mathbf{y}\|^2}{2\sigma^2}}, \qquad (7)$$

$$K(\mathbf{x},\mathbf{y}) = tanh(\kappa \mathbf{x} \cdot \mathbf{y} - \delta). \qquad (8)$$

The formula (6-8) is polynomial kernel function; Gaussian radial basis kernel function and $tanh$ kernel function respectively. Among them, the Gaussian radial basis kernel function is the most popular one. And we adopt Gaussian radial basis function in our experiment.

Transforming the original problem to the dual form of the original problem is another feature of SVM. When the original problem is hard to solve, sometimes we can resort to its dual problem to make it easier to solve. The convex quadratic programming guarantees that we can find a global optimal solution for the problem. For neural network, it is incline to get a local optimal solution.

Our experiment data set is from MNIST [9] and to get the .mat form, the readers are referred to [10]. The experiment is implemented on matlab2011b, win32, Intel(R) Core (TM) i3-2310, CPU @ 2.10GHz 2.00GB (RAM). Confined to the memory, we chose 18,000 of all the samples[5] for our experiment, 1,800 for each of the number between 0 and 9.

For the 18,000 samples, we chose one half for training and another half for testing. For each number between 0 and 9, we all chose half for training and half for testing.

And we applied one of the most common classifier, support vector machine (SVM) [11] for classification.

In our experiment, the architecture of the deep auto-encoder is 784-200- 100-200-784 and 784-200-50-200-784 [12]. And we chose the third layer (with 100 or 50 neurons) to classify with SVM. The activation function we applied is $tanh$. And we corrupt the original input with masking noise, and we mask the original data as follows: we implement the masking by making the elements of the input with the index as 1:80:784. Besides, we initialize the weight matrix $W$ with each element in the following range [12]:

$$\left(-\frac{\sqrt{6}}{\sqrt{d_v + d_h}}, +\frac{\sqrt{6}}{\sqrt{d_v + d_h}}\right).$$

And we initialize the bias vector $b$ and $c$ with $0$.

For clarity, we show the architecture we use in our experiment (Fig. 3.).

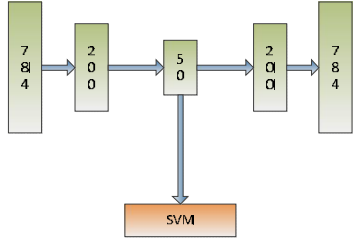

**Fig. 3.** The architecture of our model for 784-200-50-200-784, and we use the layer of "50" for classification with SVM.

The experiment result is shown in the following table (Table 1 and Table 2). And we can see that our method performs best, beating AE, DAE and CAE.

---

[5] The MNIST database of handwritten digits has a training set of 60,000 examples, and a test set of 10,000 examples.

**Table 1.** The experiment result of SVM classifying the feature extracted by AE, DAE, CAE and CDAE. The architecture is 784-200-100-200-784.

| Method | Accuracy |
|--------|----------|
| AE     | 92.42%   |
| DAE    | 92.51%   |
| CAE    | 93.11%   |
| CDAE   | 93.31%   |

**Table 2.** The experiment result of SVM classifying the feature extracted by AE, DAE, CAE and CDAE. The architecture is 784-200-50-200-784.

| Method | Accuracy |
|--------|----------|
| AE     | 93.12%   |
| DAE    | 93.28%   |
| CAE    | 93.31%   |
| CDAE   | 93.77%   |

## 4 Conclusion

In this paper, we proposed CDAE, a novel improved auto-encoder by combining CAE and DAE. CAE can make less neuron in the hidden layer active to learn more meaningful, abstract and robust features for classification, while DAE is robust to the input by first corrupting the original data and then minimizing the reconstruction loss function. Our proposed method CDAE combines the advantages of both CAE and DAE.
The experiment (classification with SVM) shows that CDAE outperforms both CAE and DAE, which also shows the effective of CDAE.
Since CDAE performed better on MNIST, we'll try CDAE on other problems in future.

## Acknowledgement

The authors thank Yuan-fang Ren for helpful discussion.